\documentclass[twoside]{article}

\usepackage[accepted]{aistats2023}
\usepackage{comment}
\usepackage[utf8]{inputenc}
\usepackage{graphicx}
\usepackage{amsmath}
\usepackage{natbib}
\usepackage{xcolor}
\usepackage{setspace}
\usepackage[colorlinks,citecolor=blue,urlcolor=blue,filecolor=blue,backref=page]{hyperref}
\usepackage{cleveref}
\usepackage{bm}
\usepackage{bbm}
\usepackage{amsfonts}
\usepackage{amssymb}
\usepackage{multicol}

\usepackage{algorithmicx}
\usepackage[noend]{algpseudocode}
\usepackage{algorithm}

\algrenewcommand\alglinenumber[1]{{\sf\scriptsize{\color{brown} #1}}}
\algrenewcommand\algorithmicrequire{\textbf{Input:}}

\newcommand{\npart}{N_\text{part}}
\newcommand{\nproc}{N_\text{proc}}
\newcommand{\nppp}{N_\text{ppp}}

\begin{document}

\begin{titlepage}
    \begin{description}
        \item[Article title:] 
        A Partitioned Sparse Variational Gaussian Process for Fast, Distributed Spatial Modeling
        \item[Running head:]
        Partitioned Sparse Variational Gaussian Process
        \item[Authors] \ 
        \begin{itemize}
            \item Michael J. \underline{Grosskopf} \\
            Statistical Sciences Group \\
            Los Alamos National Laboratory
            \item Kellin \underline{Rumsey} \\
            Statistical Sciences Group \\
            Los Alamos National Laboratory
            \item Ayan  \underline{Biswas} \\
            Information Sciences Group \\
            Los Alamos National Laboratory
            \item Earl \underline{Lawrence} \\
            Statistical Sciences Group \\
            Los Alamos National Laboratory
        \end{itemize}
        \item[Corresponding author:] Michael J Grosskopf, mikegros@lanl.gov
        \item[Keywords:] in situ analysis, climate, parallel computing
    \end{description}

\end{titlepage}

\twocolumn[

\aistatstitle{A Partitioned Sparse Variational Gaussian Process for Fast, Distributed Spatial Modeling}

\aistatsauthor{Michael Grosskopf \And Kellin Rumsey \And Ayan Biswas \And  Earl Lawrence }

\aistatsaddress{Los Alamos National Laboratory} ]

\begin{abstract}
The next generation of Department of Energy supercomputers will be capable of exascale computation. For these machines, far more computation will be possible than that which can be saved to disk. As a result, users will be unable to rely on post-hoc access to data for uncertainty quantification and other statistical analyses and there will be an urgent need for sophisticated machine learning algorithms which can be trained in situ. Algorithms deployed in this setting must be highly scalable, memory efficient and capable of handling data which is distributed across nodes as spatially contiguous partitions. One suitable approach involves fitting a sparse variational Gaussian process (SVGP) model independently and in parallel to each spatial partition. The resulting model is scalable, efficient and generally accurate, but produces the undesirable effect of constructing discontinuous response surfaces due to the disagreement between neighboring models at their shared boundary. In this paper, we extend this idea by allowing for a small amount of communication between neighboring spatial partitions which encourages better alignment of the local models, leading to smoother spatial predictions and a better fit in general. Due to our decentralized communication scheme, the proposed extension remains highly scalable and adds very little overhead in terms of computation (and none, in terms of memory). We demonstrate this Partitioned SVGP (PSVGP) approach for the Energy Exascale Earth System Model (E3SM) and compare the results to the independent SVGP case. 
\end{abstract}


\section{Introduction}
\label{sec:intro}
The next generation of Department of Energy supercomputers will be capable of more than $10^{18}$ floating point operations per second, an exaFLOP. In these exascale machines, the computing power has outpaced the I/O capability. That is, they can compute far more than can be saved to disk \citep{cappello2009toward}. As a result, users can no longer rely on post-hoc access to data for uncertainty quantification and other analysis tasks. Instead, users will need methods for {\em in situ} analysis in which data is analyzed in real time as the simulation runs \citep{oldfield2014}.

Many of these machines' applications will be simulations of spatial fields evolving in time. These include simulations of physics experiments, cosmology, space weather, and climate. Thus, in situ methods for spatial models will be a basic building block of many analyses \citep{grosskopf2021situ, rumsey2022hierarchical}. There will be a number of challenges in this regime \citep{shalf2010exascale}. First, the data will be quite large, so the computational and memory requirements of the model and method will need to scale well. In the in situ setting, we need estimation to be relatively fast or the spatial modeling becomes the simulation bottleneck. Second, the data will be distributed across nodes in the supercomputer with each node containing a spatially contiguous block of data. Communication is time-consuming, so we will not be able to communicate large amounts of data between nodes, but fitting spatial models to only local data can result in discontinuous prediction at node boundaries. Third, because of the increased resolution of such simulations there will likely be local behavior or small-scale features that we would like to capture.

In this paper, we develop an approach for in situ spatial modeling based on sparse variational Gaussian processes (SVGPs) \citep{hensman2013gaussian, hensman2015scalable}, requiring local models to predict local and neighborhood data to improve consistency in prediction between local models. These approaches can handle large data and be estimated quickly and can be scalably trained using decentralized, point-to-point communication from neighboring nodes. The fitting is mostly local so small-scale behavior can be captured. The occasional communication is decentralized and lightweight, but enough to encourage smoothness at the boundaries. We develop the approach in the context of the Energy Exascale Earth System Model (E3SM), the Department of Energy's flagship climate model \citep{golaz2019doe, caldwell2019doe, leung2020introduction}.

\section{Background}
\label{sec:sgp}

\subsection{Gaussian Processes}
Gaussian processes (GPs) are a state-of-the-art approach for spatial modeling and a popular model for estimation of an unknown function, $f(x)$ from a limited set of input-output pairs. The observations of the output of this function may be noisy:
\begin{align}
\begin{split}
\bm y &= f(\bm x) + \boldsymbol \epsilon, ~ \boldsymbol \epsilon \sim N(0, \beta^{-1})\\
\bm x &= [x_1, \dots, x_n].
\end{split}
\end{align}
The GP describes a probability distribution on functions over an input space with elements $x$
\[
f(x) \sim \text{GP}(\mu(x), \Sigma(x, x^\prime))
\]
defined by the mean function $\mu(x)$ and the covariance function $\Sigma(x, x^\prime)$. The covariance function gives the covariance between function values at any two points in the input space, $x$ and $x^\prime$. Its form determines the supported function space \citep{rasmussen2006gaussian}. The covariance function is often parameterized by a set of ``hyperparameters" -- a correlation length for each dimension and a process variance --  which can be inferred from the data. In addition, a ``nugget" or diagonal noise covariance matrix may be added to handle data measured with stochastic ``error". For notation purposes, this noise term will be considered part of the covariance function and its scale estimated as a hyperparameter. 

Given the observed input-output pairs $\bm x, \bm y$, we can compute the conditional distribution for any new point in the input space.
\begin{align}
\begin{split}
& f(\bm x^\ast) \mid \bm x, \bm y \sim N(\nu(\bm x^\ast), \Psi(\bm x^\ast, \bm x^\ast) ) \\
& \nu(\bm x^\ast) = \mu(\bm x^\ast) + \Sigma(\bm x^\ast, \bm x)^\intercal \Sigma(\bm x, \bm x)^{-1} \left( \bm y - \mu(\bm x)\right) \\
& \Psi(\bm x^\ast, \bm x^\ast) = \Sigma(\bm x^\ast, \bm x^\ast) \\
&\quad\quad\quad\quad\quad- \Sigma(\bm x^\ast, \bm x)^\intercal \Sigma(\bm x, \bm x)^{-1} \Sigma(\bm x^\ast, \bm x). 
\label{eq:gp_conditioning}
\end{split}
\end{align}
%

The GP posterior mean, $\nu(\bm x^\ast)$, gives the new predicted values for the output of the function at any location $\bm x^\ast$, while the posterior covariance, $\Psi(\bm x^\ast, \bm x^\ast)$, communicates uncertainty in the resulting surface due to the finite set of observations. 

Given the matrix computations above, GPs are intractable for large data with computational costs scaling as $\mathcal{O}(n^3)$ and memory requirements scaling as $\mathcal{O}(n^2)$. This limitation has led to wide research into different approaches; see \cite{liu2020gaussian} for a review of approaches to GPs with big data. 


\begin{figure}[t]
	\centering
	\includegraphics[width=0.45\textwidth]{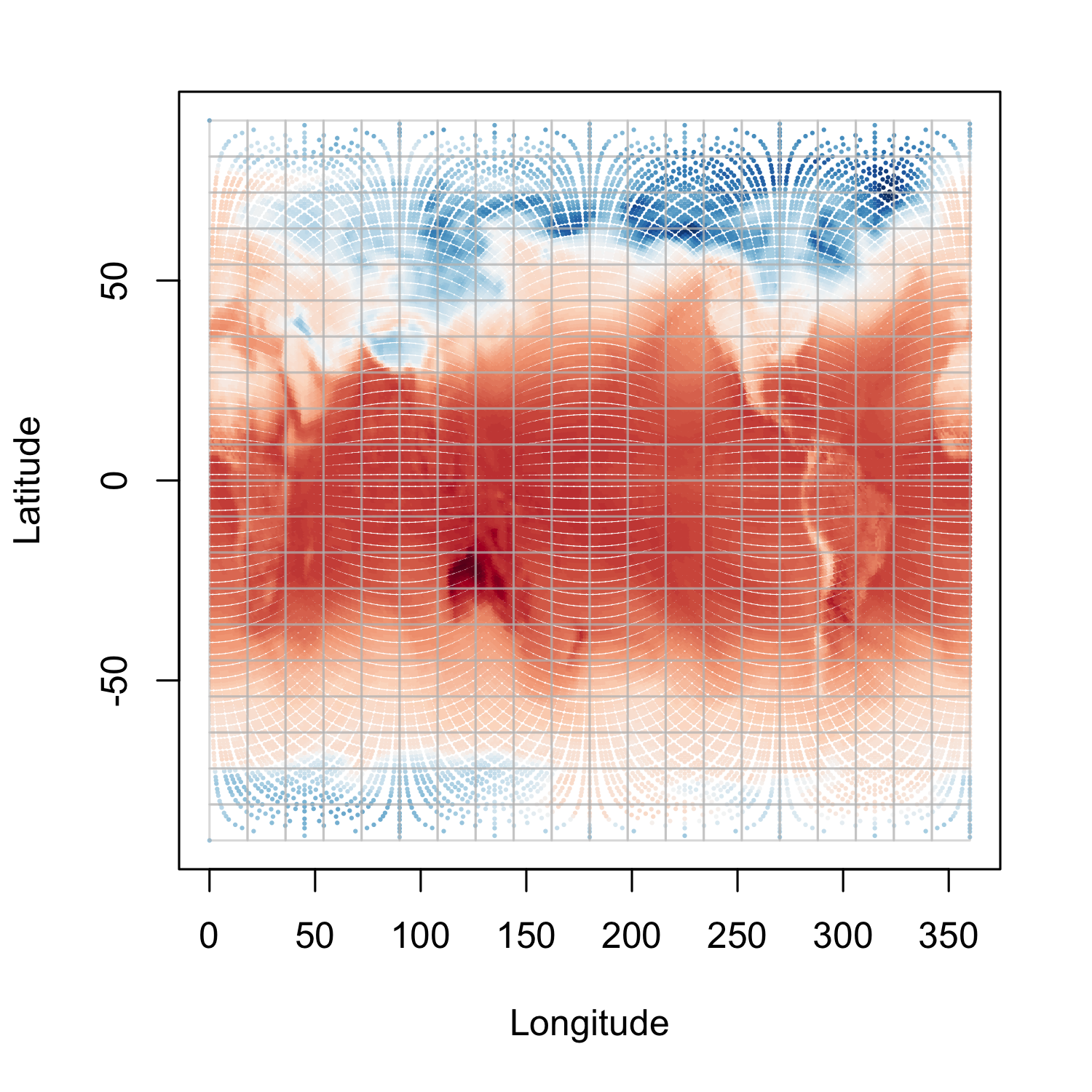}
	\caption{This image displays the output of the E3SM climate simulation for a single time slice. The gray grid represents part of a hypothetical partitioning of the domain into $\npart = 400$ contiguous data partitions.}
	\label{fig:e3sm_full}
\end{figure}

\subsection{Stochastic Variational Gaussian Process}
One common approach is a class of sparse approximations that assume that the observations are conditionally independent, given a set of $m \ll n$ {\it inducing points} \citep{titsias2009variational,hensman2013gaussian,quinonero2005unifying}. These inducing points are then optimized to obtain an optimal approximation to the full GP that scales better with data size. 
Early inducing point sparse GPs optimized the inducing point locations along with the GP hyperparameters using the marginal likelihood of the Gaussian process \citep{quinonero2005unifying}. Often the inducing point value was marginalized out, making for an easier optimization problem, with a $\mathcal{O}(nm^2)$ computational cost scaling.  \cite{titsias2009variational} proposed the inducing point approach as a form of variational inference on the full Gaussian process.

\cite{hensman2013gaussian} further innovated on the sparse variational Gaussian process model to improve the scaling by introducing a variational loss function that factored into the sum over independent terms for each observation. This allowed the use of stochastic gradient decent (SGD), improving the scalability of the sparse GP to very large data problems. 
The key to factoring the loss function was to introduce a variational approximating distribution on the output value of the inducing points. This increases the number of model parameters, in exchange for significantly improved scaling. The approximating distribution for the $m$ inducing points is the full-rank, multivariate normal distribution $q(\bm u) = N(\bm u |\bm m_\star, \bm S_\star)$, where $\bm m_\star$ and  $\bm S_\star$ are variational parameters to be optimized.  The resulting evidence lower bound (ELBO) for the \cite{hensman2013gaussian} approach, which we refer to here as SVGP, is given by
\begin{equation}
\label{eq:elbo}
    \begin{aligned}
    \text{ELBO}&(\bm\phi|\bm x, \bm y) =  \sum_{i=1}^n\ell(x_i, y_i, \bm\phi), \text{ where } \\
    \ell(x_i, y_i, \bm\phi) &\stackrel{\triangle}{=} \bigg\{\log N\left(y_i|\bm k_i^\intercal \bm K_{mm}^{-1}\bm m_\star, \beta^{-1}\right) \\ 
    & - \frac{1}{2}\left(\beta \tilde k_{ii} + \text{tr}(\bm S_\star\bm\Lambda_i)\right)  \\
    & - \frac{1}{n}\mathbb E_{\bm u}\left[\log\left(\frac{ N(\bm  u|\bm m_\star, \bm S_\star)}{p(\bm u)}\right) \right]\bigg\}, 
    \end{aligned}
\end{equation}
where $\bm \phi = (\bm m_{\star}, \bm S_{\star}, \bm z_{\star}, \bm \kappa, \beta)$ for notational compactness and $\bm\Lambda_i = \beta\left(\bm K_{mm}^{-1}\bm k_i\right)\left(\bm K_{mm}^{-1}\bm k_i\right)^\intercal$.
The inducing point locations and the covariance function parameters, $(\bm z_\star, \bm \kappa)$, are needed to construct $\bm k_i = \bm k_i(\bm x, \bm z_\star, \bm \kappa)$, $\bm K_{mm} = \bm K_{mm}(\bm z_\star, \bm \kappa)$ and $\tilde k_{ii} = \tilde k_{ii}(\bm x, \bm z_\star, \bm \kappa)$, and $\beta$ is the precision of the iid Gaussian noise vector (see \cite{hensman2013gaussian} and \cite{hensman2015scalable} for details). Optimization of this function with SGD scales as $\mathcal{O}(m^3)$, independently of the full data size $n$.

Because this approach 
explicitly learns the input location and output mean and covariance for the $m$ inducing points, it is ideal for the proposed in situ inference model. The inducing point parameters act as a parsimonious summary of the inference model while maintaining the flexibility and high predictive capability. 


\subsection{Other Related Approaches}
Other approaches for fitting fast, approximate GPs are not suitable for the in situ context described in \cref{sec:intro}. There are a number of ``local'' approaches avoid fitting a global model to the complete data such as local approximate GPs \citep{gramacy2015local}, treed GPs \citep{gramacy2008bayesian} and hierarchical mixture-of-experts models \citep{ng2014hierarchical}. However, these methods still require storage and access to the complete data. Similarly, sparse kernel approximations \citep{gneiting2002compactly, moran2020fast} lead to fast algorithms by constructing an approximation to the full covariance, but still require the complete data. Subset of data approximations \citep{chalupka2013framework} are potentially appealing for the in situ setting, but the summary provided is typically inferior to that of the SVGP \citep{liu2020gaussian}.

Additionally, there is recent work on distributed and hierarchical GP approximations which provide advantages over full-rank GPs, but are still difficult to apply in situ. \cite{park2010hierarchical} present a hierarchical GP with improved training cost, but prohibitive memory requirements. \cite{hoang2016estimating} describe a parallel, distributed GP where the observation error is treated as a realization of a $M$-order Gaussian Markov random process. However, their inducing points are shared amongst the data partitions. Work by \cite{gal2014distributed} is similar. Distributed approaches for multi-fidelity simulations \citep{fox2012multiresolution, lee2017hierarchically} are not applicable in our setting. The distributed hierarchical SVGP proposed by \cite{rumsey2022hierarchical} is suitable for in situ inference, but it can be outperformed by simpler methods when the number of inducing points is large. 

\section{Local Independent SVGPs}
\label{sec:ISVGP}

The SVGP model can be modified to handle the contiguously distributed data and to better capture small-scale structures in the output by simply fitting independent SVGP models locally to each data partition. This approach, which was successfully employed in situ for the E3SM climate simulations \citep{grosskopf2021situ}, is trivially parallelizable and accomplishes most of the goals discussed in \cref{sec:intro}. 

We assume that the data is split amongst $\npart$ partitions, so that the data for partition $j$ is given by $\bm d_j = (\bm x_j, \bm y_j)$ where $\bm x_j = [x_{j1}, \ldots x_{jn_j}], x_{ji} \in \mathbb R^d$ and $\bm y_j = [y_{j1}, \ldots, y_{jn_j}], y_{ji} \in \mathbb R$. If the number of processors available for computation is $\nproc$, then each processor is responsible for fitting and storing the local SVGP model corresponding to $\nppp = \npart/\nproc$ different partitions. The ELBO for the local SVGP model corresponding to partition $j$ is given by 
\begin{equation}
    \label{eq:elbo_isvgp}
    \text{ELBO}_j(\bm\phi_j) = \text{ELBO}(\bm\phi_j |\bm x_j, \bm y_j),
\end{equation}
where 
$\text{ELBO}(\cdot|\cdot, \cdot)$ is defined in \cref{eq:elbo}. These local ELBOs can be optimized efficiently, in parallel, with SGD. 

Although this approach, referred to hereafter as ISVGP, is well-suited for in situ analysis and scales excellently, it can lead to a highly discontinuous response surface. This occurs because there is no mechanism for enforcing models on neighboring partitions to agree on their predictions at their shared boundary. 

\begin{figure}[h]
	\centering
	\includegraphics[width=0.4\textwidth]{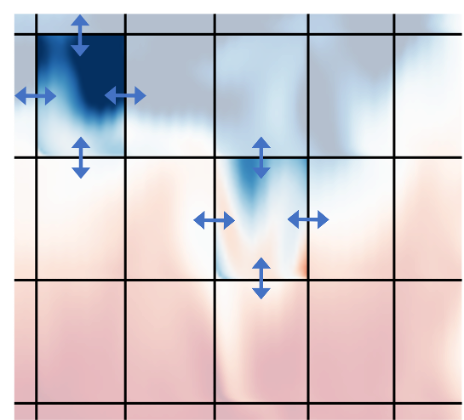}
	\caption{For each local data partition, the PSVGP uses communication of data across a local neighborhood to improve smoothness across boundaries. All MPI communication is handled by point-to-point message passing between blocks, avoiding bottlenecks from centralized communication}
	\label{fig:diagram}
\end{figure}

\section{Partitioned Sparse Variational Gaussian Process (PSVGP)}
\label{sec:psvgp}

In this section, we propose a decentralized approach to fitting sparse GP models in situ to induce smoothness compared to ISVGP, at the boundaries between partitions. The intuition is simple -- rather than have each GP fit data only within its partition, each local GP is required to have some predictive capability for data on neighboring partitions. By training local models using overlapping data, we encourage better alignment of the predictive surface at the boundary between partitions. While this approach does require communication between neighbors, this is lightweight, limited and can be controlled with a tuning parameter.

\subsection{Description of PSVGP}
\label{sec:describe}

Discontinuities in the ISVGP predictive surface occur only at the boundaries between neighboring partitions, and thus PSVGP encourages better alignment by allowing neighbors to occasionally share data. We define the {\it neighborhood set} of $j$ as 
\begin{equation}
\label{eq:neighborhood}
    \mathcal N_j = \{k \ | \ \text{partitions $j$ and $k$ share a boundary}\}
\end{equation}
and we say that $j$ and $k$ are neighbors if $k \in \mathcal N_j$. 
For notational convenience, we will take the convention that partition $j$ is always a neighbor of itself ($j \in \mathcal N_j$). 
In the fully independent SVGP algorithm (ISVGP), the local model for partition $j$ is an SVGP with data $\bm d_j$. In our proposed approach (PSVGP), the local model for partition $j$ ---an SVGP with parameters $\bm\phi_j$ and $m \ll n_j$ inducing points ---is trained using its own data as well as the data corresponding to each of the neighboring partitions, 
\begin{equation}
    {\bm D_j} = \bigcup_{k \in \mathcal N_j}\bm d_k.
\end{equation}
Since we are primarily interested in fitting this model with a large number of processors ($\nppp \approx 1$), we will need to deal with the fact that data for neighboring partitions will often be stored on different processors. Since communication across processors can be costly, special care must be taken to maintain an efficient implementation with good scaling in $\npart$. In PSVGP, the full ELBO for the $j^{th}$ local model is
\begin{equation}
\label{eq:elbo_psvgp}
    \text{ELBO}_j(\bm\phi_j | {\bm D_j}) = \sum_{k \in \mathcal N_j}\sum_{i=1}^{n_k}\ell(x_{ki}, y_{ki}, \bm\phi_j),
\end{equation}
where $\ell()$ is defined in \cref{eq:elbo}.

\subsection{Stochastic Optimization with Minimal, Decentralized Communication}
\label{sec:optimize}

Following \cite{hensman2013gaussian}, we seek to optimize \cref{eq:elbo_psvgp} using stochastic gradient descent. If we naively use a standard stochastic mini-batch approach, we are likely to sample points from each partition in $\mathcal N_j$, which would require full communication between every partition and all of its neighbors at every iteration. We can improve on the performance of standard mini-batch SGD by carefully selecting a cheap-to-evaluate stochastic function $U_j({\bm D}_j, \bm\phi_j)$ which can be computed using minimal communication. If $U_j$ is selected such that $\mathbb E_{{\bm D}_j}(U_j) = \nabla_{\bm\phi_j} \text{ELBO}_j$, then stochastic gradient descent can be used to maximize $U_j$ rather than $\nabla\text{ELBO}_j$, and the algorithm will still converge \citep{bottou2003stochastic, robbins1951stochastic}. The PSVGP optimization scheme sets the random gradient $U_j$ as 
\begin{equation}
\label{eq:elbo_psvgp_est}
\begin{aligned}
    &U_j({\bm D}_j, \bm\phi_j | k^\prime, \mathcal I^\prime) = \frac{n_{\text{eff},j} }{B}\sum_{i \in \mathcal I_{k^\prime}}\nabla\ell(x_{k^\prime i}, y_{k^\prime i}, \bm\phi_j) \\[1.2ex]
    &\mathcal I^\prime | k^\prime \sim U\left(\left\{\mathcal I \ \vert \ \mathcal I \subset \{1, \ldots, n_{k^\prime}\}, \lvert \mathcal I \rvert = B\right\}\right) \\[1.2ex]
    &\mathbb P(k^\prime = k) = \frac{n_k \mathbbm{1}(k\in \mathcal N_j)}{n_{\text{eff},j}},
\end{aligned}
\end{equation}
where $n_{\text{eff},j} \&= \sum_{k \in \mathcal N_j}n_k$ and $B$ is the mini-batch size. In words, we first select a partition $k^\prime$ from the neighborhood of partition $j$ with sampling weights proportional to the number of observations in each partition (recall that $j$ is a neighbor of itself). Next, we sample $B$ observations from partition $k^\prime$ uniformly at random and without replacement, and we use this mini-batch to obtain an unbiased estimate for the gradient of $\text{ELBO}_j(\bm\phi_j|{\bm D_j})$. The advantage of this approach relative to the naive scheme is that communication is only required if (i) $k^\prime \neq j$ and (ii) if $k^\prime$ and $j$ are handled by different processors. Even when communication between processors is required, partition $j$ communicates with at most one of its neighbors during each iteration. 

Since all of the necessary communication is point-to-point, we avoid bottlenecks associated with centralized communication. The optimization scheme, as described above, can be easily implemented using a Message Passing Interface such as Open MPI \citep{gabriel2004open} and we use the Adam optimizer \citep{kingma2014adam} to efficiently update the variational parameters $\bm\phi_j, (j=1,\ldots, \npart)$. 


\begin{figure*}[t]
	\centering
	\includegraphics[width=0.95\textwidth]{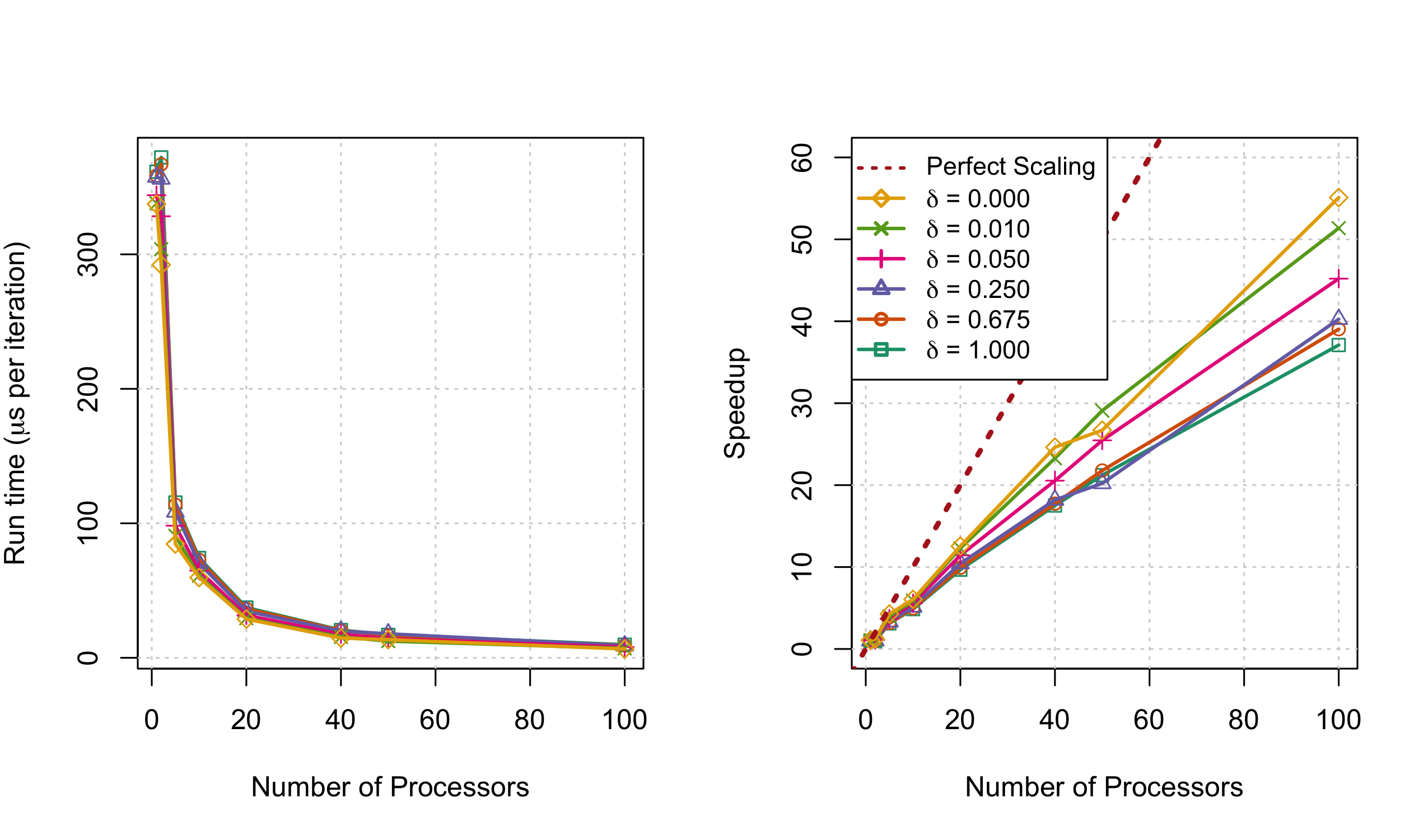}
	\caption{Runtime and scaling results PSVGP on E3SM with $\npart = 400$ partitions and $m=5$ inducing points per partition. The number of processors is usually selected so that $\nppp = \npart/\nproc$ is an integer, so that the work is well balanced across processors. The weak scaling (fixed problem size) of PSVGP is excellent, although some efficiency is lost as $\delta$ decreases. }
	\label{fig:scaling}
\end{figure*}

\subsection{Interpolating Between PSVGP and ISVGP}
\label{sec:interp}

We can reasonably expect that the response surface constructed by PSVGP will be smoother than the surface constructed by ISVGP, in the sense that neighboring local models will be in better agreement at their shared boundaries. On the other hand, PSVGP trains the local models over a larger spatial region compared to ISVGP (by a factor of $\lvert\mathcal N_j\rvert$). This suggests that PSVGP may need more inducing points to achieve accuracy which is comparable to the independent case. 

For a fixed number of inducing points $m$, it is useful to consider this trade-off between smoothness at the boundary and overall accuracy, and we posit that the ideal choice of model may lay somewhere between these two extremes. We note that our description of PSVGP above becomes equivalent to ISVGP by simply changing $n_{\text{eff},j} = n_j$ and $\mathbb P(k^\prime = k) = \mathbbm{1}(k = j)$ in \cref{eq:elbo_psvgp_est}, so that the mini-batch is always taken from the central partition. We generalize this idea by substituting 
\begin{equation}
\label{eq:delta}
\begin{aligned}
    \mathbb P(k^\prime = k) &= \begin{cases}
    \frac{n_k}{n_{\text{eff},j}}, & k = j \\[1.2ex]
    \frac{\delta n_j\mathbbm{1}(k\in \mathcal N_j)}{n_{\text{eff},j}}, & k \neq j,
    \end{cases} \\[1.2ex]
    n_{\text{eff}, j} &= \sum_{k\in \mathcal N_j}\delta n_k\mathbbm{1}(k \neq j) + n_k\mathbbm{1}(k = j)
\end{aligned}
\end{equation}
into \cref{eq:elbo_psvgp_est}, where $\delta \in (0, 1)$ is a parameter which allows us to transition continuously between ISVGP ($\delta = 0$) and PSVGP ($\delta = 1$). When $\delta$ is between $0$ and $1$, the algorithm behaves like standard PSVGP except that the optimization is biased towards sampling more frequently from the central partition, leading to smoother predictions across the boundaries while still maximizing the usefulness of the inducing points for overall predictive accuracy. In the special case where the data partitions are balanced and form a regular grid (see \cref{fig:e3sm_full}), the transformation $1 - 2d\delta/(2d+1)$ gives the proportion of the time that an interior partition will take a mini-batch from itself.
For smaller values of $\delta$, the algorithm will be more efficient, because communication is less frequent. \Cref{fig:scaling} shows, however, that PSVGP scales well even for large values of $\delta$ and is only marginally slower than ISVGP ($\delta = 0$) due to the decentralized communication of PSVGP and the modified SGD algorithm.

\section{E3SM Results}

\begin{figure*}[t]
	\centering
	\includegraphics[width=0.95\textwidth]{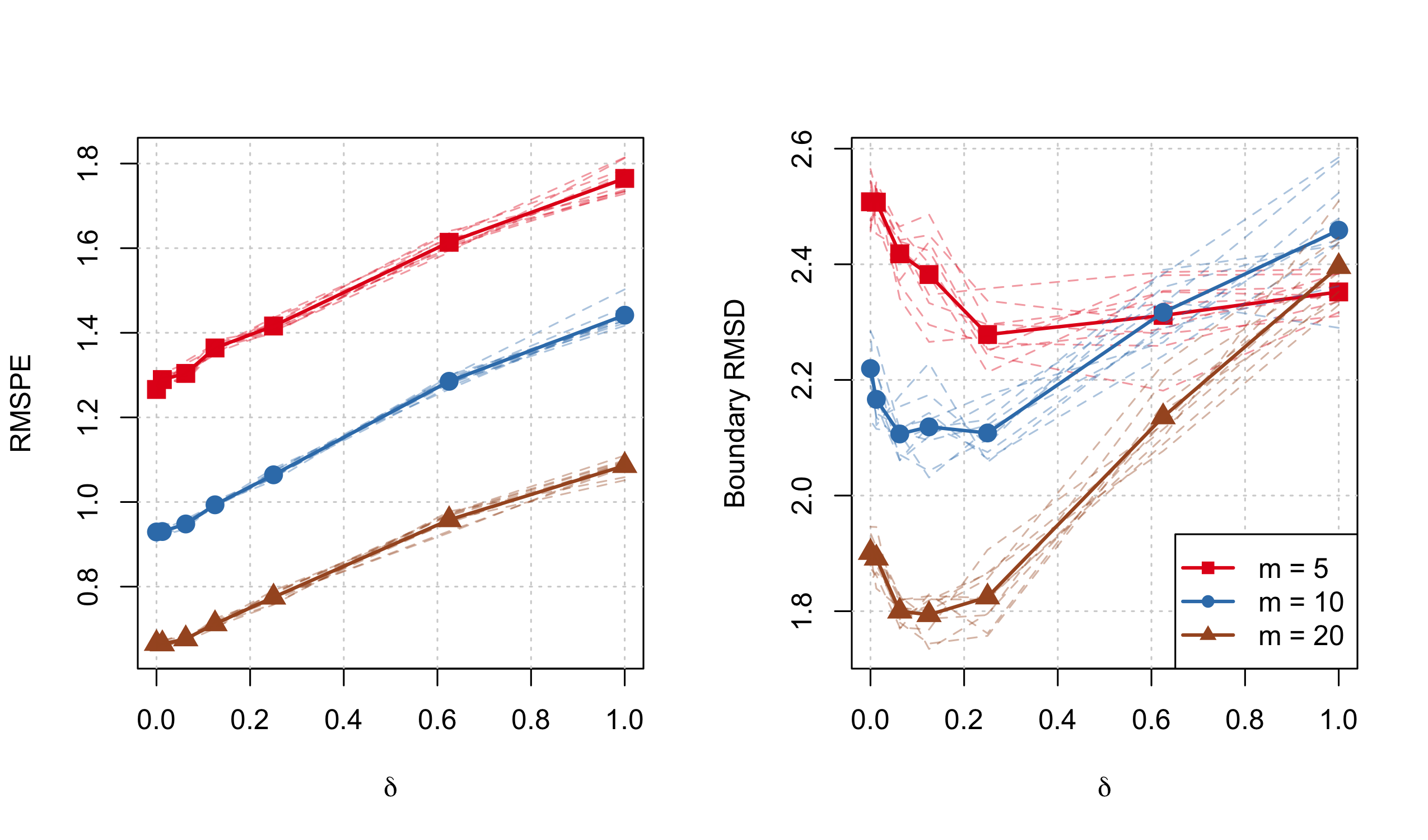}
	\caption{In-sample RMSPE (left) and smoothness at the boundary (right) is displayed as a function of $\delta$ for $m=5,10$ and $20$. Points and solid lines indicate the average across ten replications. Overall accuracy monotonically decreases with $\delta$ which is expected because more inducing points are needed for larger $\delta$. Optimally smooth spatial predictions are obtained for value of $\delta$ between $0$ and $1$. These results indicate that a value of $\delta \approx 0.15$ may be the ideal choice. }
	\label{fig:results}
\end{figure*}

In this section, we fit distributed spatial models to a single time slice of an E3SM climate simulation (see \cref{fig:e3sm_full}) consisting of $48,602$ observations across the earth. The observations are partitioned into a $20\times 20$ grid of $\npart = 400$ unbalanced partitions, which corresponds to a medium-resolution partitioning of E3SM for the purposes of testing and algorithm  development. Each partition contains between $8$ and $222$ observations with a median of $150$ observations per partition. The partitions near the poles generally have fewer observations, and thus fitting local models for latitudes near $\pm 90^\circ$ can be challenging. 

Since our spatial models are meant to be run in situ, alongside the E3SM simulation, fitting a spatial model must be fast relative to the cost of the simulator which takes around $1$ second per time step \citep{grosskopf2021situ, golaz2019doe}. To remain fast, we investigate the behavior of PSVGP using $m=5, 10$ and $20$ inducing points per partition, noting that  \cite{grosskopf2017generalized} showed that $m=5$ was a suitable choice for E3SM. The runtime of PSVGP is displayed in the left panel of \cref{fig:scaling} for $m=5$ and various $\delta \in [0,1]$, indicating that SGD can be performed for about $100$ to $150$ iterations in the same duration as a single E3SM time step (when $\nppp = 4)$. We assess the overall fit by reporting the RMSPE of each model over all observations and smoothness at the boundary is approximated by taking the root mean square difference (RMSD) between the predicted values of neighboring local models at $17,556$ locations equally spaced along the boundaries between partitions.

We compare the performance of PSVGP spatial models for a variety of $\delta$ values between $0$ and $1$. As seen in the left panel of \cref{fig:results}, the model demonstrates the best overall predictive performance when $\delta = 0$ (ISVGP), because the utility of each inducing points is locally maximized. When $\delta > 0$, the model needs more inducing points to obtain similar accuracy although (i) the loss in performance is negligible for small $\delta$ and (ii) $\delta > 0$ can lead to substantially better smoothness at the boundary (right panel of \cref{fig:results}). For example, when $\delta = 0.125$ and $m=20$, the RMSPE for PSVGP is just $1.6\%$ higher than ISVGP while the boundary RMSD decreases by $5.3\%$. The results are similar, though less pronounced, for the $m=5$ and $m=10$ cases for which the RMSPE increases by $3.0\%$ and $2.0\%$ (respectively) while the boundary RMSD decreases by $5.1\%$ and $3.6\%$ (respectively). 

\begin{figure*}[t]
	\centering
	\includegraphics[width=0.95\textwidth]{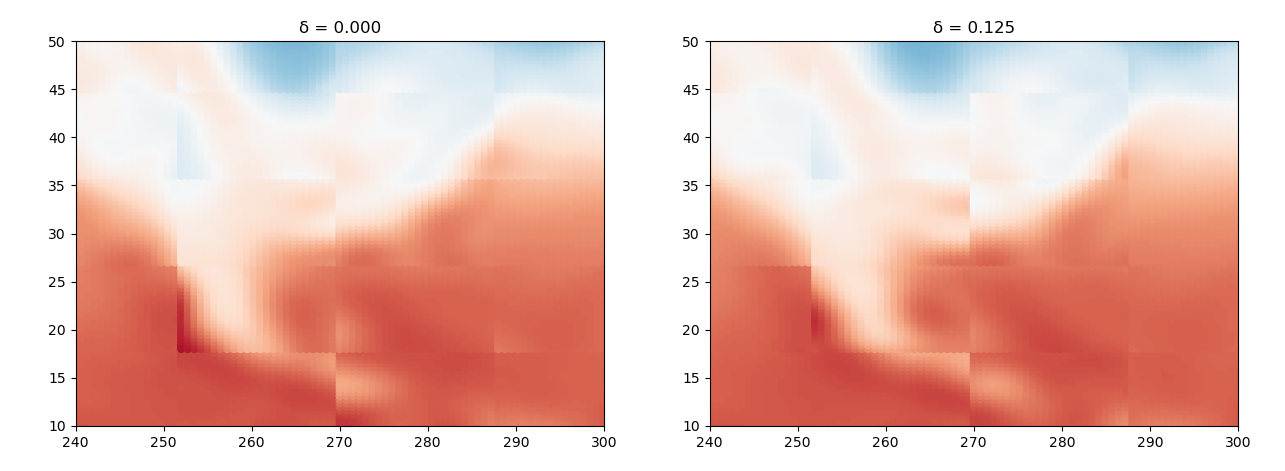}
	\caption{E3SM predictions of North America for $\delta = 0$ (ISVGP) and $\delta = 0.125$. Overall, the structure in the predictions is similar, however the PSVGP improves the smoothness along the California-Mexico coast and in the warm air mass in the northwest region.}
	\label{fig:predictions}
\end{figure*}

\Cref{fig:predictions} shows a partial view of the predictive surface for two PSVGP models with $20$ inducing points. The figure presents a close-up look at the predicted surfaces around North America using $\delta=0$ (ISVGP) and $\delta=0.125$ (the best-case result, for boundary smoothness). The model fits are generally similar, but setting $\delta > 0$ leads to improvement in several areas, most notably near the coast of California and Mexico (approximately $17^{\circ}$, $250^{\circ}$).





\section{Discussion}
The PSVGP model provides a simple solution for increasing smoothness when Gaussian processes are fit to partitioned data. The solution is highly scalable and fast enough to be run inside simulations as they are running, which means that they can be used to model and summarize data produced during complex physics simulations (e.g., E3SM) running on exascale supercomputers. The cost of improved boundary smoothness is (i) a small amount of additional computation and a slight reduction in overall RMSE. By interpolating between a pure PSVGP model and a pure ISVGP model, we can find a sweet spot for some $\delta$ (usually near $0$) where the benefit clearly outweighs the cost.

Our approach uses a relatively simple communication pattern and a novel modification of the SGD algorithm to keep the communication costs low and decentralized. In future work, we will explore more complex communication patterns that may produce better gradient estimates and understand the trade-offs with the communication burden. We will also explore post-hoc methods to increase boundary smoothness, for example based on the patchwork kriging approach of \cite{park2018patchwork}. Lastly, we have also begun studying extensions to non-Gaussian likelihoods which would allow us to model count and extreme-value data common in simulations like E3SM.

\bibliography{bibfile}

\end{document}